\documentclass[lettersize,journal]{IEEEtran}
\usepackage{amsmath,amsfonts}
\usepackage{algorithmic}
\usepackage{algorithm}
\usepackage{array}
\usepackage[caption=false,font=normalsize,labelfont=sf,textfont=sf]{subfig}
\usepackage{textcomp}
\usepackage{stfloats}
\usepackage{url}
\usepackage{verbatim}
\usepackage{caption}
\usepackage{graphicx}

\usepackage{hyperref}
\usepackage[dvipsnames]{xcolor}
\usepackage[export]{adjustbox}
\usepackage{cite}
\usepackage{tabularx}
\usepackage{pdfpages}
\begin{document}

\title{Lane Change Decision-Making through Deep Reinforcement Learning }

\author{Mukesh Ghimire$^{*}$, Malobika Roy Choudhury$^{*}$, Guna Sekhar Sai Harsha Lagudu$^{*}$
\thanks{This report is based on the implementation of ~\cite{wang2019lane}}
\thanks{$^{*}$ Email:
        {\tt\small \{mghimire,mchoud15,glagudu\}@asu.edu}} 
        
        }



\maketitle

\begin{abstract}
Due to the complexity and volatility of the traffic environment, decision-making in autonomous driving is a significantly hard problem. In this project, we use a Deep Q-Network, along with rule-based constraints to make lane-changing decision. A safe and efficient lane change behavior may be obtained by combining high-level lateral decision-making with low-level rule-based trajectory monitoring. The agent is anticipated to perform appropriate lane-change maneuvers in a real-world-like udacity simulator after training it for a total of 100 episodes. The results shows that the rule-based DQN performs better than the DQN method. The rule-based DQN achieves a safety rate of 0.8 and average speed of 47 MPH.
\end{abstract}


\section{Introduction}
\subsection{Background}
Autonomous vehicles have been gaining popularity in the recent times and have also been successfully deployed in the field. Waymo, for instance, has launched their self-driving ride sharing service, which is the first to provide fully self-driving experience to the public. While companies like Tesla and comma.ai provide some level of autonomy, they haven't been able to ship a fully autonomous vehicle to the market as of now. There are several components that make up an autonomous driving system. Primarily, these can be broken down into four different modules: perception module, decision-making module, control module, and an actuator-mechanism module~\cite{wang2019lane}. Out of these modules, this work deals with the decision-making module. Decision-making under wide variety of conditions is a hard task to generalize. Hence, it is necessary that the decision maker is robust to various conditions and can generalize or learn from its experience. As such, Reinforcement Learning (RL) comes in handy as several breakthroughs have been made in the recent years in the field of RL. 

RL, most prominently, Deep RL models have been developed that have achieved super-human performance in games like Go and Poker\cite{silver2017mastering, rebel}. In \cite{rebel}, the authors used self-play RL with search to achieve superhuman performance in heads-up no-limit Texas hold’em poker using limited domain knowledge than any existing poker AI. Advances like these have opened a wide range of potential domains where RL can be used to achieve superhuman performance. Autonomous driving also has several challenges, which currently do not have an exact solution. For instance, Waymo experienced ten instances (out of total 16 simulations) of ``same direction sideswipe'', which involved collisions during lane changing or lane merging maneuvers~\cite{waymo}.

\subsection{Related Work}
The methods to solve Autonomous Driving problems can be divided into two categories i.e Rule based and learning based algorithms. Although rule-based techniques have had some success in the past, learning-based approaches have also shown their effectiveness in recent years.\par
Many conventional solutions are based on explicitly written rules and rely on state machines to shift between specified decision behaviors. For example,the "Boss" that is trained by CMU~\cite{pomerleau1989alvinn} discusses a rule based approach, also team of researchers at Stanford~\cite{montemerlo2008junior} used  reward designs to determine trajectories.However, reliable decision has high amount of uncertainity in rule based approach.

Learning-based techniques, as a key AI technology, can deliver more advanced and safe decision-making algorithms for autonomous driving. Recently, NVIDIA~\cite{bojarski2016end} researchers trained a deep convolutional neural network (CNN) to train images directly from camera to cluster control. The trained model was able to handle the task of lane change and driving on a gravel road.

In addition to supervised learning, reinforcement learning results have significantly improved over the years. Wolf et al.~\cite{wolf2017learning} have presented a method by using DQN to drive a car in stimulated environment. Only 5 action were defined each corresponding to a different steering angle along with training images of size 48$\times$27. The reward function was calculated by taking into account the distance from lane center as well as certain auxiliary information (such as the angle error between the vehicle and the center line). Hoel et al.~\cite{hoel2018automated} used DQN to solve lane change along with vehicle speed. They used a one dimensional vector to define many components including speed, surrounding vehicles and adjacent lanes instead of using front facing images. Wang et al.~\cite{wang2019lane} used safety rate to measure the quality of their model which was calculated based on collision frequency which provided clear idea about stimulation. 
The results show that the models which considers the speed along with other factors usually outperformed the other models that did not consider the average speed of vehicle.

\subsection{Overview}
This report is an overview on the implementation of Deep Q-Learning in lane change decision-making problem with rule based constraints. This report is structured as follows. Sec.II discusses the methodology associated with the RL algorithm along with the rule-based constraint formulation. The implementation and the simulation results are discussed in Sec.III and Results are discussed in Sec.IV. Finally, the discussion and conclusion are discussed in Sec.V.

\section{Methodology}
\subsection{Markov Decision Process}
A Markov decision process (MDP) is a mathematical framework for describing decision-making in circumstances where the result is partially random and partly within the control of a decision maker. An MDP's policy satisfies the Markov property, which indicates that the conditional probability distribution of a random process's future state depends solely on the current state when the current state and all previous states are supplied ~\cite{ref1}.\par
An MDP is a 4-tuple $M = <S,A, P_{sa},R>$ ~\cite{ref1}, where
\begin{enumerate}
\item $S$ represents set of states and $s_t \in S$ which represents state at time step $t$.
\item $A$ represents set of actions and $a_t \in A$ which represents action at time step $t$.
\item $P_{sa}$ is the probability of performing action $a \in A$ in current state $s \in S$ will lead to next state.
\item $R$ is the reward function.
\end{enumerate}
At each time steps, $t = 0, 1, 2,....,$ the agent interacts with the environment and it observes the current state st S of the environment, then chooses and executes a practicable action at $a_t \in A$ depending on the state. Following the completion of the activity in the environment, the agent receives a numerical reward $r_t \in R$ and the next state $s_{t+1}$.\par
The purpose of reinforcement learning is to learn an optimum policy that maps from environmental states to agent behaviors by maximizing the cumulative reward ~\cite{ref1}. The policy at time step $t$ , indicated by $\pi t$, relates environmental states to probabilities of choosing each potential action, where $\pi^t(a|s)$ is the probability that $a_t$ = a, if $s_t = s$. Reinforcement learning normally employs a way of maximizing a total anticipated return $G$, which is a cumulative sum of immediate rewards $r$ received over the long run, to arrive at the best policy. $ G_t$ is defined at time step $t$ as
\begin{equation}
    G_t=\Sigma_{k=0}^{\infty}\gamma^kr_{t+k}
\end{equation}
where $\gamma \in (0,1] $ is the discount factor.
\subsection{State Space}
\begin{figure}[!h]
\centering
\includegraphics[width=3.4in]{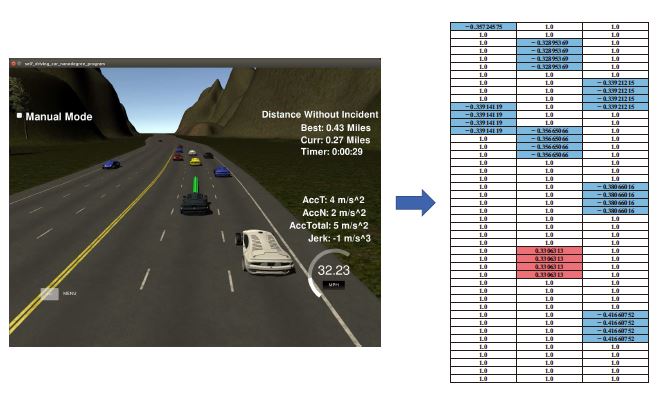}
\caption{State Representation}
\label{fig_2}
\end{figure}

The simulator provides information on the location and speed of the ego car as well as other vehicles. It comprises the $x, y$ position of the cars in map coordinates and $s, d$ position in fernet coordinates, the self-driving car's yaw angle in the map and speed in MPH, and the $x, y$ speed in m/s of the other cars. In this project, $s,d$ positions of vehicles in Frenet co-ordinates and the speed of the vehicle are used to represent the states of the vehicles. A $45 \times 3$ matrix is used to represent the state space, which corresponds to the entire traffic situation within the range of 60 meters front and 30 meters behind the ego car as shown in Fig. ~\ref{fig_2}. Each car is approximately 5.5 meters in length and each row in the matrix spans 2 meters, one car can occupy 4 cells in extreme cases. Hence, we fill the 4 cells corresponding an individual car with the car’s normalized speed. The speed is normalised with respect to the maximum speed of the vehicle in frame and minimum speed of vehicle in the frame. Here, the normalized speed of the ego car is positive while other cars are in negative. Cells corresponding to location without any cars are filled with 1s. 
\subsection{Action Space}

The RL agent (ego car) has three actions to choose from -- go to the left lane, stay in the current lane, or go to the right lane. All the other controls are handled by the low level controller. Table~\ref{tab:actions} shows the individual actions available to the agent. Actions are enumerated from 0 to 2 for each of the cases. 

\begin{table}[!h]
\caption{Action Space for the RL Agent\label{tab:actions}}
\centering
\begin{tabular}{|c|c|}
\hline
Action & Description\\
\hline
$0$ & stay in the current lane\\
\hline
$1$ & go to the right lane\\
\hline
$2$ & go to the left lane\\
\hline
\end{tabular}
\end{table}

\subsection{Reward}

The self-driving car in this project is only allowed to drive on the right 3 lanes of the highway. In case of an illegal lane change i.e., when action $2$ ($or$ $1$) is selected when the car is on the left-most ($or \; right$) lane, the car is forced to stay in the same lane with a negative reward of $r_{ch1}=-5$.  A negative reward of $r_{co}=-10$ is assigned to avoid collision. A penalty of $r_{ch2}=-3$ is given to the agent when it decides to change lane even when there is no car in front of it (invalid lane-change case).


On top of changing lanes only when required, the ego car should also drive as fast as possible (within the speed limit). To enforce this, following reward is assigned. 

\begin{equation}
r_v=\lambda(v-v_{ref})
\end{equation}
where $v$ represents ego car's average speed, $v_{ref}$ represents reference speed (25 MPH), and $\lambda$ represents normalizing coefficient which is equal to $0.04$. Our reward structure within one decision period is

\begin{equation}
    r=\left \{
    \begin{matrix}
    r_{co} && \text{collision happens}\\
    r_{ch1} && \text{illegal lane change} \\
    r_{ch2} && \text{invalid lane change} \\
    r_{v} + r_{ch3} && \text{legal lane change}\\
    r_{v} && \text{normal drive}\\
    
    \end{matrix}
    \right.
\end{equation}

When lane change happens without collision.

\subsection{Rule-Based Constraints}
We apply rule-based limitations based on the planning trajectories and the others' expected trajectories to assure absolute security of the lane change behavior. The low-level controller can anticipate the trajectories of the ego car and the adjacent cars in the desired lane depending on the action determined by the high-level decision maker.

The trajectories of adjacent cars are calculated assuming they retain their present speed and stay in the current lane. The choice made by the high-level decision maker is potentially harmful if the distance between the anticipated trajectory of the ego car and that of surrounding cars is smaller than the specified threshold value. This action gets rejected by the low-level controller and the car will continue on its current trajectory. 
\subsection{Deep Q Network} \label{dqn}

Deep Q-Learning is used to determine the optimal action for a given state. Q-Learning is an off-policy TD reinforcement learning algorithm. It evaluates a policy $\pi$ using the state-action value function $Q^{\pi}(s, a)$, which is defined as: 

\begin{equation}
    Q^{\pi} (s, a) = \mathbb{E}_{\pi}  \left[ \sum_{k=0}^{\infty} \gamma^{k}\;r_{t+k} \bigg| s_t = s, a_t = a \right] 
\end{equation}

The Q-learning algorithm tries to find the optimal state-action value function:
\begin{equation}
    Q^*(s, a) = \max\limits_{\pi} Q^\pi (s, a)
\end{equation}

$Q^*$ corresponds to the optimal policy $\pi^*$.

This value function $Q*$ follows the Bellman Optimality equation: 
\begin{equation}
    Q^* (s, a) = \mathbb{E}\left[r + \gamma \max\limits_{a' \in A} Q^* (s', a') \bigg|s, a \right]
\end{equation}

Using the optimal value function $Q^*$, the optimal policy can be determined by finding actions that maximize the value function at each state~\cite{wang2019lane}. 

\begin{figure}[!h]
\centering
\includegraphics[width=2.4in]{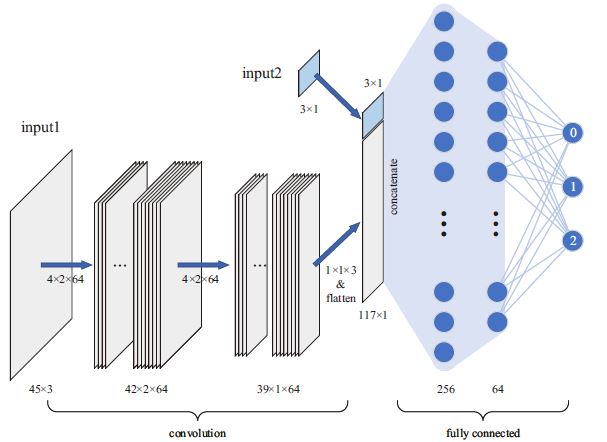}
\caption{Deep Q-Network Architecture}
\label{fig_a}
\end{figure}

Finding a Q-function is simple enough for a discrete and finite problem. However, for problems like ours with high dimensional and continuous state-space, this method becomes computationally expensive and intractable. A Deep Q-Network, as shown in Figure~\ref{fig_a}, is used to approximate the value function. Following loss function is minimized using Adam optimizer~\cite{wang2019lane}:
\begin{equation}
    L_i(\theta_i) = \mathbb{E} \left[(r+\gamma\max\limits_{a'} Q(s', a'; \theta_i^-) - Q(s, a; \theta_i))^2 \right],
\end{equation}

The input to the DQN is the state matrix along with a $3D$ vector, which is concatenated to the network after the convolution layers. The first element of the vector corresponds to the normalized speed of the RL agent, while the second and the third element are either 1 (True) or 0 (False) if there is a lane to the left or right respectively.  


\section{Implementation and Simulation} \label{simulation}
\subsection{System Requirements}
For our project we used Ubuntu
with 16 GB RAM, Intel I5 processor and 4 GB GPU. The
following software installations are required for the project:
\begin{enumerate}
\item{CMake 3.22.0}  
\item{Python-3.6}
\item{Virtual Box-6}
\item{Ubuntu.iso}
\item{CUDA 11.1.1}
\item{Tensorflow-GPU}
\item{keras}
\end{enumerate}

Also, a dedicated GPU was used for better training performance.
\subsection{Implementation}
\begin{enumerate}
\item{If not on Ubuntu, Install Virtual box version 6 or above on your system} 
\item{Download Ubuntu desktop version 16.04 or above}
\item{Now designate sufficient virtual memory and processing power to the virtual box for Ubuntu to run smoothly}
\item{Import the disk image of Ubuntu on to the Virtual Box}
\item{git clone \url{https://github.com/DRL-CASIA/Autonomous-Driving.git }}
\item{ Go to folder "term3-sim-linux" }
\item{sudo chmod u+x term3-sim for 64 bit }
\item{Ensure that cmake greater than 3.5, make greater than or equal to 4.1, gcc/g++ greater than or equal to 5.4 }
\item{Enter in the CarND-test folder and run install-ubuntu.sh to install dependencies (bash install-ubuntu.sh) }
\item{Keep in the CarND-test folder and compile: mkdir build and cd build cmake .. }
\item{Modify the last line in environment.yaml to your conda installation location, and run conda env create -f environment.yaml to create a virtual environment}
\item{Install PyCharm IDE}

\end{enumerate}

\subsection{Simulator}
\begin{figure}[!ht]
\centering
\includegraphics[width=2.5in]{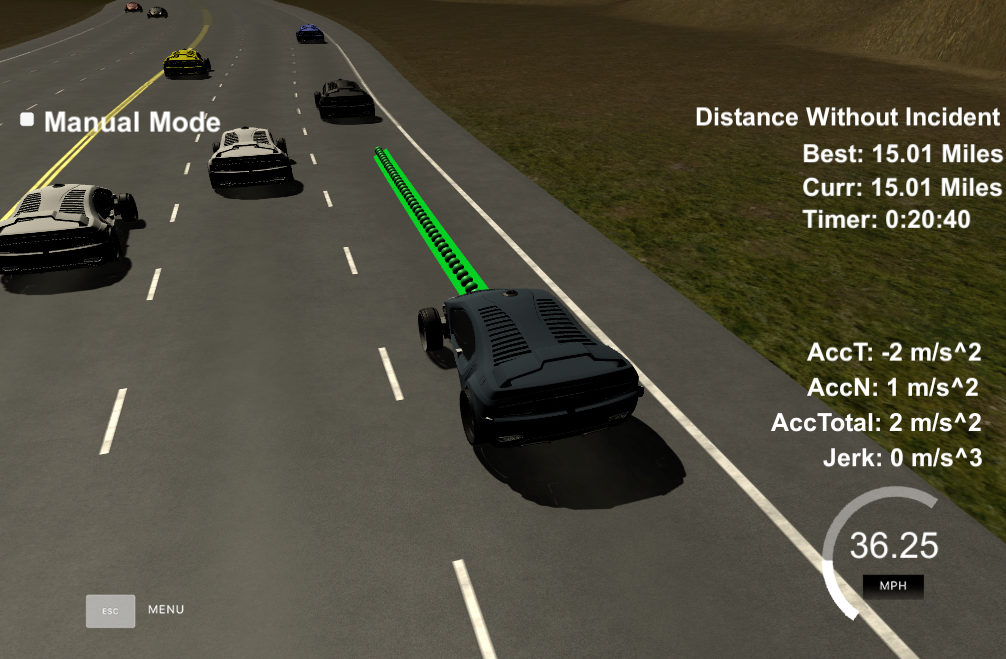}
\caption{Udacity Simulator}
\label{fig_1}
\end{figure}

The simulator used in this project is developed by Udacity~\cite{udacity}. The simulation setup consists of a 3-lane highway as shown in Fig.~\ref{fig_1}. The vehicle with the green trajectory marker is the RL agent. The simulator provides the localization and speed of all the cars in the frame. The speed limit is 50 MPH, and the goal of the RL agent is to maintain the speed limit. The simulator also provides feedback related to the performance of the car in terms of acceleration (normal, tangential, and resultant) and jerk. The requirements for the car are such that it should not experience acceleration of over  $10 m/s^2$ and jerk greater than $10 m/s^3$. The total distance of the track is 6946 meters.

\subsection{Training Details}
The Deep Q-Network model as discussed in Sec.\ref{dqn} is trained for 100 episodes. The input to the network consists of state matrix that encodes the traffic condition. The output of the network are the three state-action values (or Q-values) for each action. The action corresponding to the maximum state-action value is picked. During training the agent picks action either randomly or following the greedy policy. This phenomenon, known as the exploration-exploitation trade-off is handled by implementing a $\epsilon$-greedy policy. With $\epsilon$-greedy policy, the agent selects random action with the probability of $\epsilon$, and greedy-action with $1-\epsilon$ probability. A decaying $\epsilon$ is used with $\epsilon_{0} = 1$, $\lambda_{decay} = 0.99985$, and $\epsilon_{min} = 0.03$.   
Training steps:
\begin{enumerate}

\item{Download the updated code from github from \href{https://github.com/ghimiremukesh/Autonomous-Driving/tree/master/decision-making-CarND}{\textcolor{blue}{here}}}\footnote{\href{https://github.com/ghimiremukesh/Autonomous-Driving}{https://github.com/ghimiremukesh/Autonomous-Driving}}.
\item{Check the port and host in the code and make sure the port is free}
\item{Run train.py file following which simulator opens up}
\item{Select the simulation window and graphics quality and hit start}

\end{enumerate}


\subsection{Simulation Environment}
There are three aspects to our simulation experiment's implementation. The simulator  is the initial component, which generates environmental data and receives a predetermined path. The second component is the controller and planner, which is in charge of speed control and course planning. The DQN algorithm, which is in charge of high-level lane change decision-making, is the third component.\par
The goal of this study is to use deep reinforcement learning to make lateral lane change decisions. The low-level controller includes a rules-based speed controller and a path planner, but not the data processing. A rule-based technique is used for longitudinal speed control, while spline interpolation is used for path planning based on the specified waypoints and prospective target spots matching to the lane change decision outcome. Simultaneously, the controller serves as a lower-level modifier, allowing higher-level decisions to be revised.\par
\begin{figure}[!h]
\centering
\includegraphics[width=2.3in]{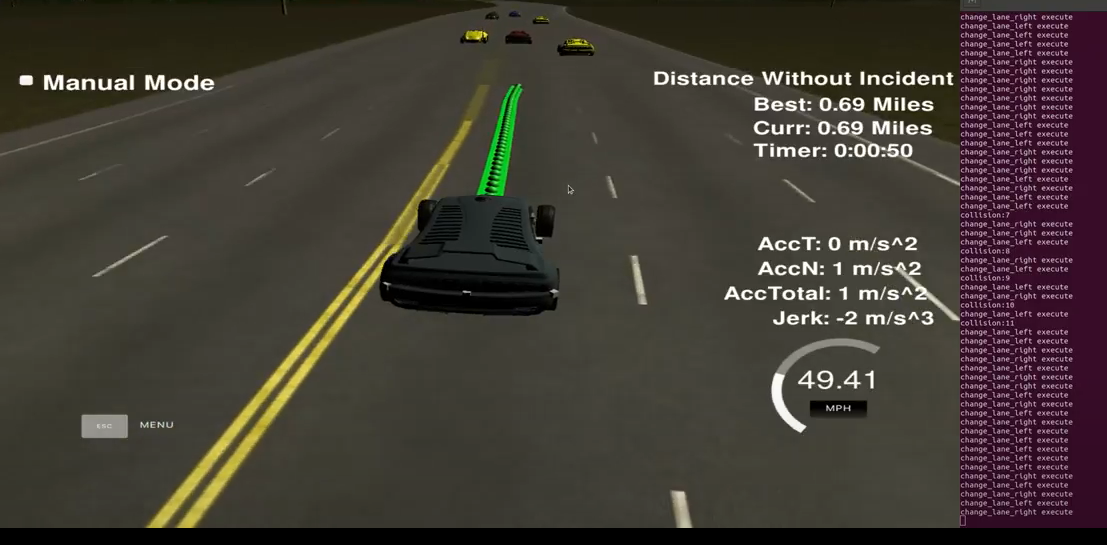}
\caption{Simulation During Training and the RL Agent's Action}
\label{train_1}
\end{figure}
\begin{figure}[!h]
\centering
\includegraphics[width=2.3in]{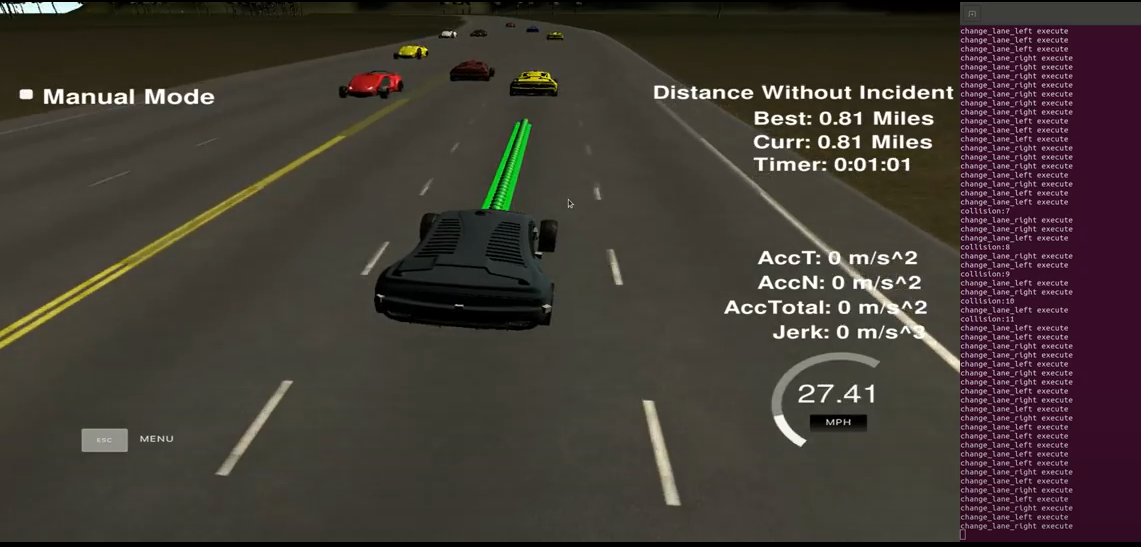}
\caption{Agent changes lane after an action is executed}
\label{train_2}
\end{figure}

When a training is launched, the simulator loads the environment and waits for control action from the RL agent while following the low-level controller. In the beginning of the training, the car always starts on the middle lane. The low-level rule-based controller maintains the motion of the car in straight line avoiding forward collision. The action picked by the RL agent (either based on the policy or randomly) triggers lane change, which is communicated to the simulator. The simulator then executes the lane change based on the action it receives. An instance of the training is shown in Fig.~\ref{train_1}. The last action executed by the agent is to go to the right lane. This action is communicated to the simulator and the car goes to the right lane as seen in Fig.~\ref{train_2}. A few time steps of training is made available \href{https://drive.google.com/drive/u/3/folders/1e9CaX4bE08x__5hfRL_mxWcqufk7Tjcj}{\textcolor{blue}{here}}.

\begin{figure}[!t]
    \centering
    \includegraphics[width=2.3in]{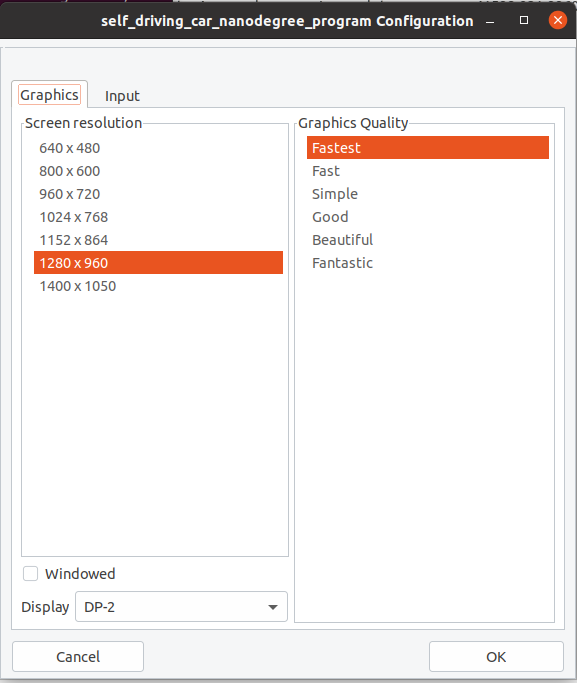}
    \caption{Initial Menu Selection in the Simulator}
    \label{fig:menu_1}
\end{figure}

\begin{figure}[!h]
    \centering
    \includegraphics[width=2.3in]{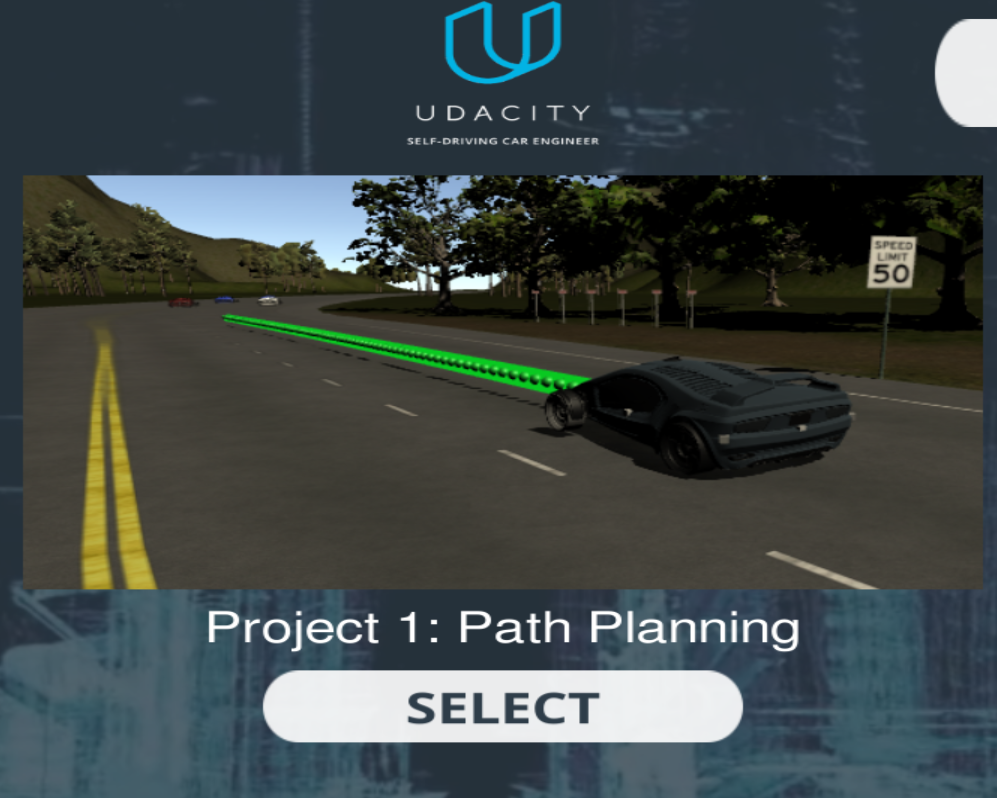}
    \caption{Final Menu Selection in the Simulator}
    \label{fig:menu_2}
\end{figure}

One of the initial challenges with the simulator was that the simulator and the terminal had to restart at the end of every episode. It required navigating through the simulator menus as shown in figures~\ref{fig:menu_1} and~\ref{fig:menu_2}. It was not feasible to physically be present to select the options when training for extended number of episodes. We use PyAutoGUI package to automate this process. We found that closing any extra window other than the IDE was smooth in terms of the package's performance. Furthermore, having a dual monitor setup would sometime lead to the simulator opening in random co-ordinates in the screen. Hence, we use single monitor for easier execution. The package can automate mouse-clicks and keyboard strokes. For mouse clicks we recorded the co-ordinates of the radio buttons that were to be clicked and passed it to PyAutoGUI for execution.

\section{Results}
\begin{figure}[!htb]
\centering
\includegraphics[width=3.15in]{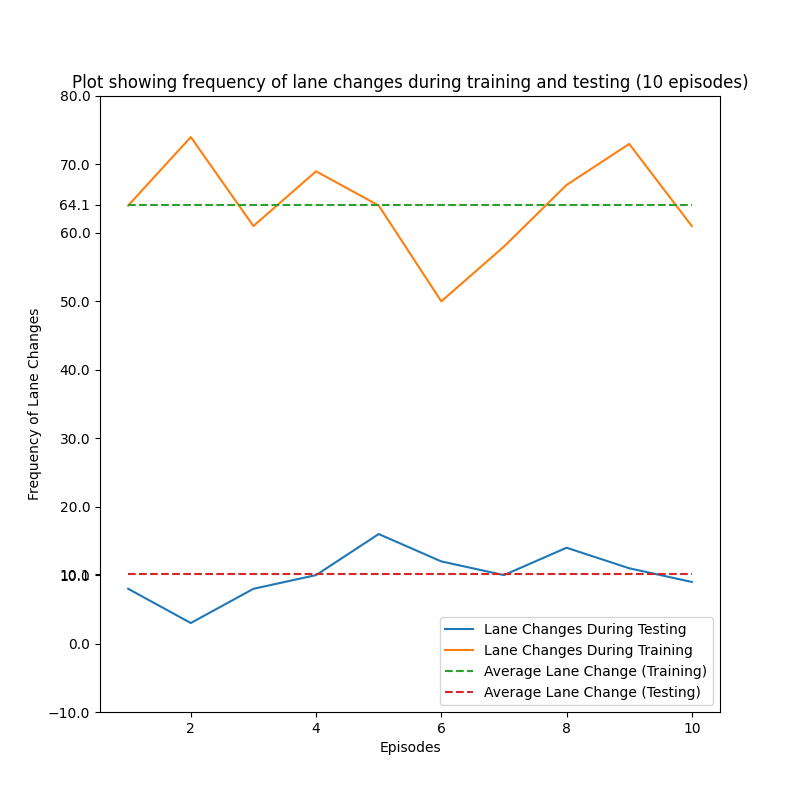}
\caption{Frequency of Lane Changes}
\label{result}
\end{figure}

The simulator operates in real time, and it takes around 6 minutes for the ego vehicle to complete a lap in the simulation environment, which is viewed as an episode in our training process. The  training program is made up of 100 episodes. When one episode is over, we restart the simulator for the next episode. Figure~\ref{result} shows the frequency of lane changes during training and testing. The figure only shows the results for first 10 episodes of training and testing. 

\begin{figure}[!htb]
\centering
\includegraphics[width=3.15in]{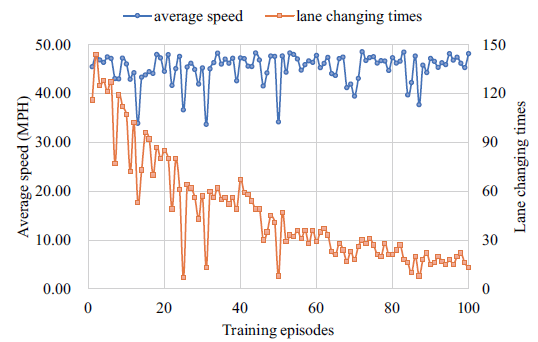}
\caption{Results from the reference paper (Lane changes during training)}
\label{result_paper}
\end{figure}

Lane changing fully depends on the traffic condition at the given time-step. From fig.~\ref{result}, we see that during training, the agent changes lanes quite often, on average about 64.1 times. After training for 100 episodes and testing the agent for 10 episodes, we see that the frequency of the lane changes drops significantly, with an average of 10 lane changes per episode. Furthermore, as the training progresses, we see a gradual decrease in lane change trend. However, due to exploration, we do see erratic behavior. The results after training show that the agent has learned to only change lanes whenever necessary. Comparing the results obtained in the reference paper (cf. ~\ref{result_paper}), we do not see such decrease in the lane changing frequency during training. They do not compare the results from training and testing. One possible theory could be that their exploration rate was low and their policy was to always pick the greedy action.  

We compare the average speed and average lane change times of the trained rule-based DQN agent to those of DQN-based policy. We run the agent through the simulation environment and then compute its average speed, average number of lane changes, and safety rate. The safety rate is defined as the ratio of test episodes without crashes to total test episodes.The results are given in Table ~\ref{results}. Here. the rule based DQN method has highest safety rate and less number of lane changes ($0.8$ and $8.80$ respectively) than the traditional DQN method ($0.2$ and $36.20$ respectively). This implies that our technique results in a more efficient and secure strategy than the others.

\begin{table}[!htb]
    \centering
    \caption{Results}
    \begin{tabular}{|c|c|c|c|}
    \hline
         Method &Average Speed&(in MPH)  & Safety rate\\
         \hline
         DQN-based policy & 46 & 36.20 & 0.2\\
         \hline
         rule-based DQN policy& 47 & 8.80 & 0.8 \\
         \hline
    \end{tabular}
    
    \label{results}
\end{table}

\section{Discussion and Conclusion} \label{Discussion}

The aim of this project was to replicate the results obtained in the referenced paper while understanding the concepts of Deep Q-Learning and its application to Reinforcement Learning in Autonomous Driving. Initially, as with most of the programming tasks, we faced several difficulties during compilation of the source code. Since the project was open-sourced we were able to easily get hands on with the source code, however, setting up the right environment was hard as there were little to no documentation on the requirements. Most of our time was invested in setting up the training/testing environment and reviewing the existing code. Furthermore, gathering resources for training was an equally hard task. As we know, training a deep neural network is computationally expensive task and especially in the context of reinforcement learning, it requires several iteration of running this expensive training task. As our application required running a simulator, training on a ``command-line only'' cluster was not possible. Hence, our only option was using one GeForce GTX 1080 GPU. 

We compared the results obtained from our training with the results shown in the referenced paper. Due to resource constraints we did not compare the results with other Reinforcement Learning Frameworks, which is the main idea of the referenced paper for this project. The results that we obtained resembled the results mentioned in the paper. However, on several iterations, we noticed that though the ego vehicle caused a ``minor'' collision, it resulted in series of collisions behind the ego vehicle. In real-life scenario, this is a catastrophic accident which must be highly penalized. Modifying reward function could enable further improvements in this method. One possible future work could be conducting an ablation study on how formulating different reward functions changes the performance of the agent.

\bibliography{ref}

\begin{thebibliography}{10}
\providecommand{\url}[1]{#1}
\csname url@samestyle\endcsname
\providecommand{\newblock}{\relax}
\providecommand{\bibinfo}[2]{#2}
\providecommand{\BIBentrySTDinterwordspacing}{\spaceskip=0pt\relax}
\providecommand{\BIBentryALTinterwordstretchfactor}{4}
\providecommand{\BIBentryALTinterwordspacing}{\spaceskip=\fontdimen2\font plus
\BIBentryALTinterwordstretchfactor\fontdimen3\font minus
  \fontdimen4\font\relax}
\providecommand{\BIBforeignlanguage}[2]{{%
\expandafter\ifx\csname l@#1\endcsname\relax
\typeout{** WARNING: IEEEtran.bst: No hyphenation pattern has been}%
\typeout{** loaded for the language `#1'. Using the pattern for}%
\typeout{** the default language instead.}%
\else
\language=\csname l@#1\endcsname
\fi
#2}}
\providecommand{\BIBdecl}{\relax}
\BIBdecl

\bibitem{wang2019lane}
J.~Wang, Q.~Zhang, D.~Zhao, and Y.~Chen, ``Lane change decision-making through
  deep reinforcement learning with rule-based constraints,'' 2019.

\bibitem{silver2017mastering}
D.~Silver, J.~Schrittwieser, K.~Simonyan, I.~Antonoglou, A.~Huang, A.~Guez,
  T.~Hubert, L.~Baker, M.~Lai, A.~Bolton \emph{et~al.}, ``Mastering the game of
  go without human knowledge,'' \emph{nature}, vol. 550, no. 7676, pp.
  354--359, 2017.

\bibitem{rebel}
N.~Brown, A.~Bakhtin, A.~Lerer, and Q.~Gong, ``Combining deep reinforcement
  learning and search for imperfect-information games,'' \emph{arXiv preprint
  arXiv:2007.13544}, 2020.

\bibitem{waymo}
M.~Schwall, T.~Daniel, T.~Victor, F.~Favaro, and H.~Hohnhold, ``Waymo public
  road safety performance data,'' 2020.

\bibitem{pomerleau1989alvinn}
D.~A. Pomerleau, ``Alvinn: An autonomous land vehicle in a neural network,''
  1989.

\bibitem{montemerlo2008junior}
M.~Montemerlo, J.~Becker, S.~Bhat, H.~Dahlkamp, D.~Dolgov, S.~Ettinger,
  D.~Haehnel, T.~Hilden, G.~Hoffmann, B.~Huhnke \emph{et~al.}, ``Junior: The
  stanford entry in the urban challenge,'' \emph{Journal of field Robotics},
  vol.~25, no.~9, pp. 569--597, 2008.

\bibitem{bojarski2016end}
M.~Bojarski, D.~Del~Testa, D.~Dworakowski, B.~Firner, B.~Flepp, P.~Goyal, L.~D.
  Jackel, M.~Monfort, U.~Muller, J.~Zhang \emph{et~al.}, ``End to end learning
  for self-driving cars,'' 2016.

\bibitem{wolf2017learning}
P.~Wolf, C.~Hubschneider, M.~Weber, A.~Bauer, J.~H{\"a}rtl, F.~D{\"u}rr, and
  J.~M. Z{\"o}llner, ``Learning how to drive in a real world simulation with
  deep q-networks,'' IEEE, pp. 244--250, 2017.

\bibitem{hoel2018automated}
C.-J. Hoel, K.~Wolff, and L.~Laine, ``Automated speed and lane change decision
  making using deep reinforcement learning,'' IEEE, pp. 2148--2155, 2018.

\bibitem{ref1}
``Reinforcement learning: An introduction,'' 2018.

\bibitem{udacity}
Udacity, ``Carnd-path-planning-project,''
  \url{https://github.com/udacity/CarND-Path-Planning-Project}, 2019.

\end{thebibliography}
\bibliographystyle{IEEEtran}

\end{document}